\documentclass{article}
\usepackage{spconf,amsmath,graphicx}

\usepackage[dvipsnames]{xcolor}
\usepackage{tablefootnote}
\usepackage{booktabs}
\usepackage{multirow}
\usepackage{adjustbox}
\usepackage{siunitx}
\usepackage{float}
\usepackage{hyperref}
\hypersetup{
    colorlinks=true,
    linkcolor=blue,
    filecolor=magenta,
    urlcolor=cyan,
}
\usepackage{amssymb}
\usepackage{pifont}
\newcommand{\xmark}{\ding{55}}%

\newcommand{\etal}{\textit{et al.}}

\def\L{{\cal L}}

\title{RCDPT: Radar-Camera fusion Dense Prediction Transformer}
\name{Chen-Chou Lo and Patrick Vandewalle}
\address{EAVISE, PSI, Dept.\ of Electrical Engineering (ESAT), KU Leuven \\
Jan de Nayerlaan 5, 2860 Sint-Katelijne-Waver, Belgium}
\begin{document}
%

\maketitle
%


\begin{abstract}

Recently, transformer networks have outperformed traditional deep neural networks in natural language processing and show a large potential in many computer vision tasks compared to convolutional backbones. 
In the original transformer, readout tokens are used as designated vectors for aggregating information from other tokens.
However, the performance of using readout tokens in a vision transformer is limited.
Therefore, we propose a novel fusion strategy to integrate radar data into a dense prediction transformer network by reassembling camera representations with radar representations.
Instead of using readout tokens, radar representations contribute additional depth information to a monocular depth estimation model and improve performance.
We further investigate different fusion approaches that are commonly used for integrating additional modality in a dense prediction transformer network.
The experiments are conducted on the nuScenes dataset, which includes camera images, lidar, and radar data.
The results show that our proposed method yields better performance than the commonly used fusion strategies and outperforms existing convolutional depth estimation models that fuse camera images and radar.

\end{abstract}
%


\begin{figure*}
\vspace{-9mm}
  \includegraphics[width=\textwidth,height=10cm]{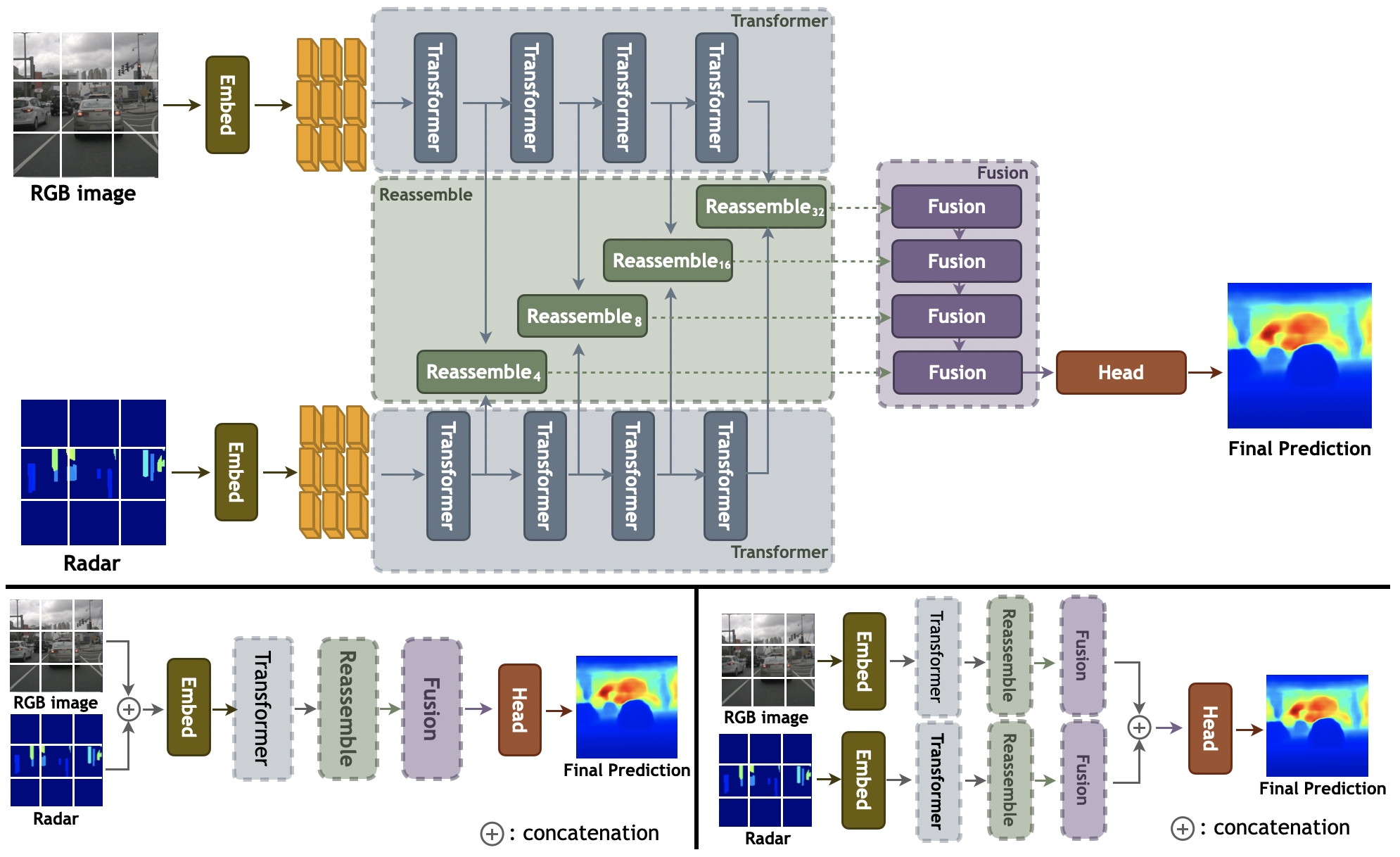}
  \caption{
  \textit{Top:} Proposed method overview. The overall structure from DPT \cite{DPT} is used. We reassemble camera image representations with radar representations instead of using the readout token in reassemble layers. 
  \textit{Bottom-left:} Illustration of an early fusion method. Camera images and radar data are concatenated at the input.
  \textit{Bottom-right:} Illustration of a late fusion method. Extracted features from camera images and radar are concatenated after fusion layers.
  }
  \label{fig:arch}
  \vspace{-2mm}
\end{figure*}


\vspace{-3mm}
\section{Introduction}
\label{sec:intro}
\vspace{-3mm}

Depth information serves as a fundamental intermediate feature in an autonomous vehicle's tasks, such as 3D object detection, recognition and segmentation.
Therefore, dense depth estimation plays a crucial role in 3D understanding scenarios.
In recent years, researchers have proposed many approaches for estimating dense depth based on monocular camera images \cite{bts,DORN}, stereo camera images \cite{PSMNet, GA-Net}, and multi-modal methods that fuse camera images and lidar data \cite{s2d,self_s2d}.
Although lidar can provide complementary information compared to camera images, the sensitivity to weather conditions and high cost still limit the commercial use of lidar as a depth sensor.
On the other hand, radar has been used as a robust depth sensor in military and commercial applications for decades because of its robustness, stability, and low cost.
Accordingly, researchers started integrating radar data as an additional guidance signal with camera-based depth estimation models on nuScenes, a sizeable autonomous driving dataset was also released including monocular camera images, radar, and lidar.

Most of the existing depth estimation models follow an encoder-decoder architecture. 
A convolutional classification network is always used as the backbone network in the encoder part to extract valuable features from input data.
Extracted features are aggregated from various stages of the encoder and merged into a final estimated depth in the decoder part.
However, convolutional backbones downsample image features into multiple scales to gain a larger receptive field while remaining within an acceptable computation. 
One of the drawbacks of these downsampling mechanisms for a depth estimation model is that the details of the original resolution will be lost. 
As a result, many proposed works focus on improving decoders by leveraging multiple scales of features at various stages \cite{bts, MSG-CHN} to recover details at a higher resolution. 
Next to convolutional backbones, vision transformers (ViT) \cite{vit} have recently been used as an alternative backbone. 
Unlike convolutional operations, the input features are split into patches and embedded into bag-of-word representations in the vision transformer backbone. 
The embedded representations maintain a fixed dimension through all stages in the vision transformer, which can preserve details and lead to a fine-grained prediction. 

In this work, we propose a novel fusion structure\footnote{Source code is available at \href{https://github.com/lochenchou/RCDPT}{https://github.com/lochenchou/RCDPT}.} to integrate radar data into a camera-based dense prediction transformer (DPT) \cite{DPT}. 
Embedded camera representations are reassembled with a readout token in DPT.
However, Ranftl \etal claimed in \cite{DPT} that the purpose of a readout token needs to be clarified, and a series of experiments were conducted to show the best way to integrate a readout token.
Instead of reassembling camera representations with a readout token as proposed in DPT, we reassemble camera representations with radar representations for radar features that could provide additional depth information.
We also investigate some commonly used fusion strategies, early and late fusion, for integrating radar data in a depth estimation model based on the structure of DPT.
The experiments are conducted on the nuScenes dataset, which includes camera images, radar, and lidar.
The results show that the proposed method yields better performance than commonly used fusion strategies.
We further demonstrate that the proposed method achieves state-of-the-art performance compared to existing convolutional depth estimation models that fuse camera images and radar.

The contributions of this work can be summarized as follows: 
(1) we investigate early and late fusion strategies to integrate radar data into a dense prediction transformer network;
(2) we proposed using radar representations to replace readout tokens in reassemble layers, which improves the overall performance.


\vspace{-5mm}
\section{Related work}
\label{sec:related_work}
\vspace{-3mm}
\subsection{Multi-Modal Depth Estimation}
\vspace{-3mm}

Based on the success of camera-based depth estimation models, many works \cite{s2d,MSG-CHN,guide_uncertainty} have further proposed adding lidar as a guidance signal because the integration of lidar can significantly improve the overall performance. 
Since radar is cheap, robust, and reliable, researchers have also worked on fusing radar data with depth estimation models. 
Lin \etal \cite{MDE_radar} firstly investigated different fusion strategies for integrating raw radar in a camera-based depth estimation model. 
Lo and Vandewalle \cite{DORN_radar} developed a preprocessing method to enlarge the depth information in radar data and fused radar in a late-fusion manner based on a deep ordinal regression network \cite{DORN}.
Lee \etal \cite{MDE_multitask} leveraged additional heads for semantic segmentation and 2D object detection to improve depth estimation performance.
Long \etal \cite{RC_PDA} proposed a radar camera association network to densify projected radar depth while resolving the uncertainties. 
These existing multi-modal depth estimation models fusing camera and radar have shown the potential of radar data to improve the performance of camera-based depth estimation models. 

\vspace{-3mm}
\subsection{Vision Transformer for Depth Estimation}
\vspace{-3mm}
The attention mechanism in transformer networks makes transformers a leading architecture in natural language processing \cite{transformer}. 
Since Dosovitskiy \etal \cite{vit} successfully utilized transformers in image recognition tasks and achieved a state-of-the-art performance, computer vision researchers have started investigating the possibilities of transformers for depth estimation tasks. 
Aditya \etal \cite{fusion_transformer} proposed a multi-modal method to integrate image and lidar representations and show the potential of a fusion approach based on a transformer backbone.
Ranftl \etal \cite{DPT} introduced a dense prediction transformer (DPT) that reassembles embedded vectors from the multi-stage of a vision transformer encoder and decodes features back to a pixel-wise depth prediction via a CNN decoder. 
Zhao \etal \cite{monovit} employed plain convolutional layers to obtain local features and combine them with global features from transformer layers.
Chang \etal \cite{ASTransformer} proposed an attention-based up-sample block to compensate for the texture features and an attention supervision mechanism to provide further guidance for transformer layers.
Han \etal \cite{TransDSSL} proposed an attention-based decoder module to enhance fine details while keeping global context from the vision transformer.


\begin{figure*}[ht]
\vspace{-9mm}
  \includegraphics[width=\textwidth]{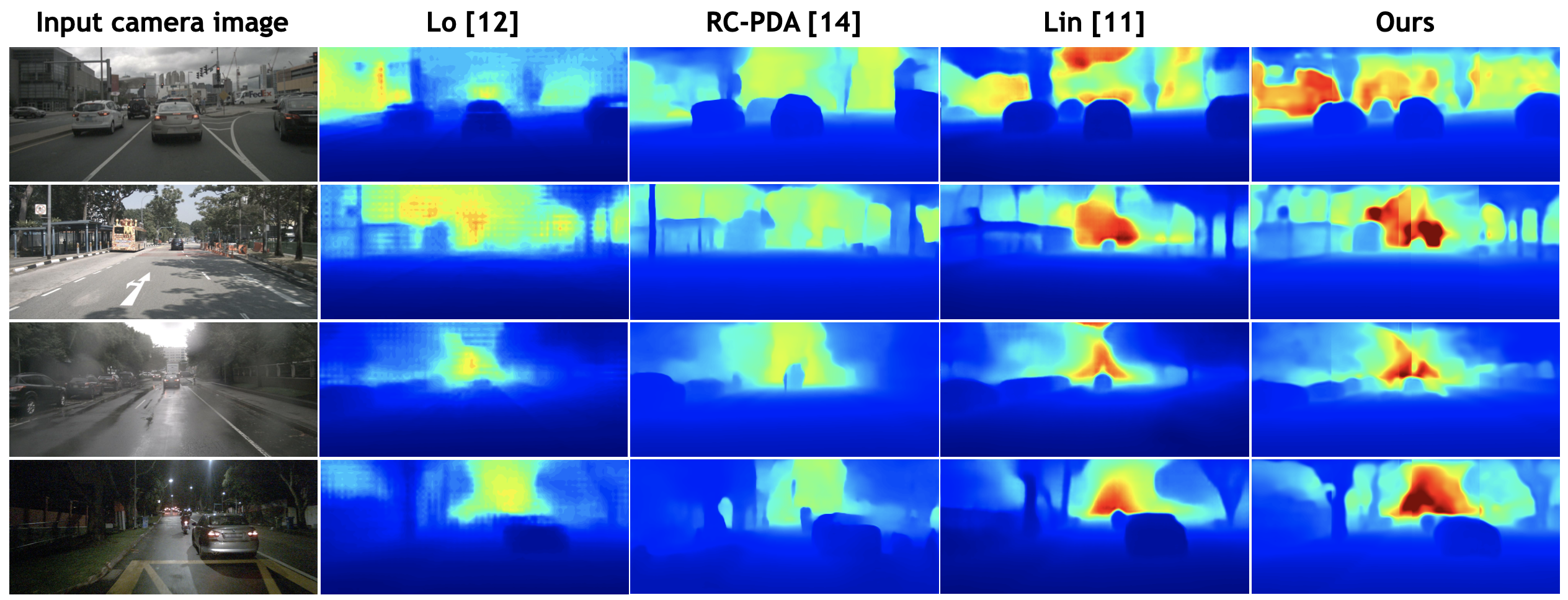}
  \caption{
  Sample results from our proposed method and previous works on nuScenes dataset. Predictions are generated based on released pre-trained weights from each project's github. Depth range is from $0-80m$.
  }
  \label{fig:result}
  \vspace{-3mm}
\end{figure*}


\vspace{-5mm}
\section{Method}
\label{sec:method}
\vspace{-3mm}

This section introduces the proposed radar-camera fusion dense prediction transformer, RCDPT. 
The overall structure from DPT [1] is utilized, and the multi-modal reassemble layers are proposed. 
An overview of the proposed method, early and late fusion structures are depicted in Fig. \ref{fig:arch}.

\vspace{-4mm}
\subsection{Radar-Camera Dense Prediction Transformer}
\vspace{-3mm}

Firstly, input images and radar are transformed into tokens by processing all non-overlapping square patches of size $p^2$ pixels and then embedded individually by a linear projection. 
The embedding is a set of $t^0=\{t^0_0,...,t^0_{N_p-1}\}, t^0_n \in \mathbb{R}^D $ tokens, where $D$ is the feature dimension and $N_p=\frac{HW}{p^2}$ for an input of size $H \times W$. 
The embedded tokens are then transformed through $L$ transformer layers, resulting in $t^l$, where $l$ refers to the $l$-th layer in the transformer backbone.
We use ViT-Base \cite{vit} as the transformer backbone encoder, $p=16$, and $D=768$ in our experiments.

Since the purpose of the readout token is unclear in the DPT structure, we replace it by fusing image and radar tokens in the adapted reassemble layers instead. 
\vspace{-2mm}
\begin{equation}
\begin{array}{l}
Reassemble^{\hat{D}}_{s}(t_{I},t_{R})= \\
(Resample_{s} \circ Concatenate \circ Read_{proj})(t_{I},t_{R}),
\end{array}
\end{equation}
\begin{equation}
\begin{array}{l}
Read_{proj}(t_{I},t_{R}):\mathbb{R}^{N_p \times D \times 2} \mapsto \mathbb{R}^{N_p \times D} \\
=mlp(cat(t_{I},t_{R})),
\end{array}
\end{equation}
\begin{equation}
Concatenate:\mathbb{R}^{N_p \times D} \mapsto \mathbb{R}^{\frac{H}{p} \times \frac{W}{p} \times D},
\end{equation}
\begin{equation}
Resample_{s}:\mathbb{R}^{\frac{H}{p} \times \frac{W}{p} \times D} \mapsto \mathbb{R}^{\frac{H}{s} \times \frac{W}{s} \times {\hat{D}}},
\end{equation}

\noindent
where $\hat{D}$ is the output feature dimension, $s$ denotes the output size ratio of the feature maps compared to the input size, and $t_{I},t_{R}$ are transformed tokens of the input image and radar respectively. 
The tokens from both modalities are concatenated and mapped to dimension $D$ using a linear layer in $Read_{proj}$.
The projected features are reshaped into an image-like feature map of size $\frac{H}{p} \times \frac{W}{p} \times D$ in the $Concatenate$ stage. 
The $Resample_s$ operation further re-scale features into size $\frac{H}{s} \times \frac{W}{s} \times {\hat{D}}$.
We use tokens from layers $l=\{3,6,9,12 \}$ of ViT-Base, and the dimension $\hat{D}=256$.

After reassembling layers, we follow the fusion layers and depth head proposed in \cite{DPT}.
The reassembled features are combined from consecutive stages and upsampled in each fusion layer.
Finally, a depth estimation head is used to produce a dense prediction at the exact resolution of input images and radar, as shown in Fig. \ref{fig:arch}.

\vspace{-4mm}
\subsection{Early and Late Fusion}
\vspace{-3mm}

We also compare our proposed approach to early and late fusion methods using a transformer network as the backbone encoder.
The illustrations in the bottom-left and bottom-right in Fig. \ref{fig:arch} are the early and late fusion methods, respectively.
For the early fusion method, we concatenate camera images and radar as the input feature and go through the same structure as in DPT to get the final prediction.
For the late fusion method, camera images and radar have a separate transformer, reassemble, and fusion branches.
The extracted features from both branches are concatenated after the fusion stage, and a depth head is used to obtain the final prediction.

\begin{table*}
\vspace{-9mm}
\centering
\caption{
Quantitative results comparing the proposed method with existing models on the nuScenes dataset.
\textit{Top rows:} baseline models using monocular images.
\textit{Middle rows:} existing camera and radar fusion methods based on convolutional backbones.
\textit{Bottom rows:} proposed RCDPT, early and late fusion approaches.
}
\label{tab:result}
\begin{tabular}{lcccccc}
\hline\hline
\multicolumn{1}{c}{\textbf{Method}} & \textbf{Radar Format} & \textbf{$\delta < 1.25$ $\uparrow$} & \textbf{$\delta < 1.25^2$ $\uparrow$} & \textbf{$\delta < 1.25^3$ $\uparrow$} & \textbf{RMSE $\downarrow$} & \textbf{AbsRel $\downarrow$} \\ 
\hline
DORN \cite{DORN} & \xmark & 0.872 & 0.952 & 0.978 & 5.382 & 0.117 \\
S2D \cite{s2d} & \xmark & 0.862 & 0.948 & 0.976 & 5.613 & 0.126 \\
DPT \cite{DPT} & \xmark & 0.886 & 0.957 & 0.980 & 5.244 & 0.106 \\ 
\hline
S2D \cite{s2d} & Raw & 0.876 & 0.949 & 0.974 & 5.628 & 0.115 \\
Lo (single-stage) \cite{DORN_radar} & Height-extend & 0.889 & \textbf{0.961} & \textbf{0.984} & 5.191 & 0.109 \\
Lo (two-stage) \cite{DORN_radar} & Height-extend & 0.890 & 0.960 & 0.983 & 5.260 & 0.108 \\
Lin (single-stage) \cite{MDE_radar} & Raw & 0.884 & 0.953 & 0.977 & 5.409 & 0.112 \\
Lin (two-stage) \cite{MDE_radar} & Raw & \textbf{0.901} & 0.958 & 0.978 & 5.180 & 0.100 \\
Lee \cite{MDE_multitask} & Raw & 0.895 & 0.958 & 0.978 & 5.209 & 0.104 \\ 
RC-PDA \cite{RC_PDA} & MER & 0.830 & 0.917 & 0.956 & 6.942 & 0.128 \\
\hline
DPT-Early & MER & 0.892 & 0.956 & 0.978 & 5.401 & 0.099 \\
DPT-Late & MER & 0.888 & 0.958 & 0.981 & 5.207 & 0.104 \\
RCDPT-Reassemble & MER & \textbf{0.901} & \textbf{0.961} & 0.981 & \textbf{5.165} & \textbf{0.095} \\
		
\hline\hline
\end{tabular}

\vspace{-3mm}
\end{table*}


\vspace{-4mm}
\subsection{Training Objective}
\vspace{-3mm}

We apply $\L_{1}$ loss as our training objective and add an additional smoothness term to the loss to retain sharp transitions at object edges. 
\vspace{-2mm}
\begin{equation}
\L_{smooth} = |\nabla_{u}(Y)|e^{-\nabla_{u}(I)} + |\nabla_{v}(Y)|e^{-\nabla_{v}(I)},
\end{equation}
\begin{equation}
\L_{total}=\omega_1*\L_{1} + \omega_2*\L_{smooth},
\end{equation}

\noindent
where $Y$ is the final prediction, $I$ is the input camera image, and $\nabla_{u}$ and $\nabla_{v}$ denote the gradient in $u$ and $v$ directions respectively.
We use $\omega_1$ = 1 and $\omega_2$ = 0.1 in our experiments.


\vspace{-4mm}
\section{Experiments}
\label{sec:experiments}
\vspace{-3mm}

\subsection{Implementation Details}
\vspace{-3mm}

All the experiments are implemented in PyTorch and trained on a Tesla V100 GPU. 
We closely follow the implementation details used by Ranftl \etal \cite{DPT}.
We conducted our experiments on the nuScenes dataset \cite{nuScenes} for most previous works are also based on this dataset. 
We use the official 700 training and 150 validation scenes as our training and validation targets. 
The transformer backbone is initialized with pre-trained weights from \cite{vit}.
A polynomial decay with a starting learning rate of 0.0001 and a power rate of 0.9 is used as the learning strategy.
The batch size is set to 4, and momentum and weight decay are set to 0.9 and 0.0005, respectively. 
The input camera images are normalized with the mean and standard deviation from ImageNet.
During training, both input camera images and radar depth are cropped into a size of $384 \times 384$, and we further use data augmentation for camera images to improve the robustness as follows: gamma contrast in range (0.9, 1.1); brightness adjustment in range (0.9, 1.1); color adjustment in range (0.9, 1.1); horizontal flipping with 0.5 probability. 
For radar data, we use the multi-channel enhanced radar (MER) proposed by \cite{RC_PDA} with a threshold of 0.5.
This results in 25319 training and 5409 validation samples.
Models are updated with the loss calculated against ground truth sparse lidar accumulated with 4 previous frames and trained for 60 epochs.

\vspace{-5mm}
\subsection{Evaluation Metrics}
\vspace{-3mm}
We evaluate the results with the following metrics. 

\noindent
{\bf Threshold Accuracy ($\delta$$_n$):} 
$\%$ of $Y$ s.t. 

$
max(\frac{Y^*}{Y},\frac{Y}{Y^*}) = \delta_n < 1.25^n 
$

\noindent
{\bf Root Mean Square Error (RMSE):}

$
\sqrt{\frac{1}{N} \sum_{i=1}^{N} \left \| Y-Y^* \right \|^2_2 }
$

\noindent
{\bf Absolute Relative Error (AbsRel):}

$
\frac{1}{N} \sum_{i=1}^{N} \frac{\left| Y-Y^* \right|}{\left| Y \right|} 
$

\noindent
where $i$ is pixels and $N$ is the total pixel number. $Y$ and $Y^*$ are the dense prediction and the target depth respectively.

\vspace{-4mm}
\subsection{Results on nuScenes}
\vspace{-3mm}
The quantitative results of the proposed method and previous works on the nuScenes dataset with depth range $<80m$ are summarized in Table \ref{tab:result}.
We refer the interested reader to \cite{MDE_radar}, \cite{DORN_radar} and \cite{RC_PDA} for details of raw radar, height-extended radar and MER respectively.
In the top three rows of Table \ref{tab:result}, we can see that the DPT model outperforms DORN and S2D in experiments without radar data and even matches some metrics to the performance of radar-camera fusion models.
The proposed RCDPT shows the state-of-the-art performance when compared to existing works. 
The result illustrates the power of a transformer backbone compared to existing convolutional networks.
This can be attributed to the fine-grained and global representations of the transformer network.
Fig. \ref{fig:result} shows qualitative results of the proposed method compared with previous works.
It shows the same trend: our method yields better details not only in the main objects but also in the surrounding environments.
Although some evaluated metrics are comparable to previous works, the proposed method shows better and smoother outputs in the qualitative comparison.

\vspace{-4mm}
\subsection{Ablation Study}
\vspace{-3mm}
To further verify the effectiveness of our proposed method, we compare the performance of the proposed method to both early and late fusion approaches with the same DPT-based network architecture.
The results are summarized in the bottom three rows in Table \ref{tab:result}.
Although early and late fusion improves the performance on most metrics when compared with the baseline DPT model, the improvements are limited.
It is clear that the proposed reassemble method outperforms early and late fusion in all metrics. 
This result further indicates that instead of using a readout token, reassembling with radar provides a better alternative.
Regarding the inference time, the baseline DPT is 10 fps while our proposed method is 9 fps.
The additional transformer layers for radar cause additional computational cost, and the remaining layers have the same structure as the baseline DPT.


\vspace{-6mm}
\section{Conclusion}
\label{sec:conclusion}
\vspace{-3mm}

We have proposed a radar-camera fusion depth estimation model based on DPT that uses a transformer as the backbone.
Instead of using readout tokens, the extracted features from the image and radar are merged effectively at the proposed reassemble layer.
Furthermore, we also investigated commonly used early and late fusion methods to demonstrate the effectiveness of our proposed method. 
The results of the experiments show that our proposed method outperforms existing works that use a convolutional backbone.
The ablation study also depicted that the proposed reassemble strategy yields better performance compared to early and late fusion on the same transformer structure.

\noindent\textbf{Acknowledgements}: This work was funded by a KU Leuven-Taiwan MOE Scholarship and Internal Funds KU Leuven.


\bibliographystyle{IEEEbib}
\vspace{-2mm}
\bibliography{refs}

\end{document}